\documentclass[11pt]{article}

\usepackage[final]{acl}
\usepackage{times}
\usepackage{latexsym}
\usepackage[T1]{fontenc}
\usepackage[utf8]{inputenc}
\usepackage{microtype}
\usepackage{inconsolata}
\usepackage{graphicx}
\usepackage{booktabs}
\usepackage{amsmath,amssymb}
\usepackage{subcaption}
\usepackage{url}
\usepackage{multirow}
\usepackage{tabularx}
\usepackage{makecell}
\usepackage{array}
\usepackage{xcolor}
\usepackage{pifont}

\newcommand{\sici}{\textsc{SICI}}
\newcommand{\favg}{$F_{\mathrm{avg}}$}

\title{\sici: A Semantic-Pragmatic Complexity Index Reveals Regime Shifts in LLM Stance Detection}

\author{
\textbf{Fuqiang Niu}\textsuperscript{1}, 
\textbf{Bowen Zhang}\textsuperscript{2}\\
  \textsuperscript{1}School of Cyber Science and Technology, \\University of Science and Technology of China, Hefei, China\\
  \textsuperscript{2}School of Artificial Intelligence, Shenzhen Technology University, Shenzhen, China\\
  }

\begin{document}
\maketitle

\begin{abstract}
Prompt-based LLMs are increasingly used for stance detection, but harder examples are not always repaired by clearer instructions, reasoning prompts, retrieval, or debate. We introduce \sici{} (Stance Inference Complexity Index), a seven-dimensional diagnostic measure of the semantic-pragmatic burden imposed by a target--text pair. Across SemEval-2016 and VAST, \sici{} predicts LLM accuracy better than surface proxies and shows substantial cross-scorer reliability ($\alpha=0.771$). More importantly, LLM errors change regime as \sici{} increases: low-complexity examples invite over-attribution, especially \textsc{Against} predictions; intermediate examples form an unstable boundary; and high-complexity examples rapidly concentrate on \textsc{None}. This phase-transition-like structure persists across GPT-3.5, GPT-4o-mini, DeepSeek-V3, and GPT-4o, although stronger models move the boundaries. A 15-method intervention study further shows that prompting, retrieval, and debate often shift models along the attribution--abstention axis rather than removing the high-complexity bottleneck.
\end{abstract}

\section{Introduction}
\label{sec:introduction}

Stance detection asks whether a text expresses \textsc{Favor}, \textsc{Against}, or \textsc{None} toward a given target \citep{mohammad2016semeval,ALDAYEL2021102597}. With the rise of large language models (LLMs), recent work increasingly treats stance detection as prompt-based inference: a model receives the text, target, label definitions, and sometimes demonstrations or reasoning instructions, then outputs one of the stance labels. This paradigm has naturally led to richer elicitation strategies, including zero-shot and few-shot prompting, chain-of-thought reasoning, evidence-oriented prompting, retrieval augmentation, and multi-agent discussion or voting.

These strategies are motivated by a plausible assumption: if an LLM already possesses the relevant linguistic and world knowledge, then clearer instructions, more examples, explicit reasoning, or additional context should elicit better stance judgments. The empirical picture for hard stance examples is less settled. Prior work has explored chain-of-thought, counterfactual, and knowledge-enhanced prompting for stance detection \citep{weinzierl-harabagiu-2024-tree,taranukhin-etal-2024-stance,zhang-etal-2024-llm-driven}, while recent work also studies multi-path reasoning, implicit target augmentation, and stereotype-sensitive LLM evaluation \citep{zhang-etal-2025-mprf,ji-etal-2025-llm-driven,dubreuil-etal-2025-stereotypes}. These studies suggest that additional reasoning or external information does not automatically yield reliable gains when the stance is implicit or the target-related evidence is weak. This raises a basic question: what limits prompt-based LLM stance detection on hard examples?

We argue that this limitation cannot be understood solely as a problem of how the model is prompted. Stance examples differ substantially in the inferential burden they impose. Some texts explicitly mention the target and state a direct attitude. Others discuss an adjacent issue, rely on pragmatic implication or background knowledge, contain sentiment whose target is ambiguous, or simply lack enough evidence to support a stance judgment~\citep{zhang2022sentiment}. Hard examples are therefore not merely harder versions of the same label-selection problem: they may require qualitatively different amounts of semantic-pragmatic inference before a stance label can be licensed.

To study this source of variation, we introduce \sici{}, the \textbf{Stance Inference Complexity Index}. \sici{} combines seven semantic-pragmatic dimensions: target visibility, scope alignment, pragmatic implicitness, knowledge requirement, context dependence, label ambiguity, and polarity--stance gap. It is not intended to replace standard metrics such as accuracy, macro-F1, or the SemEval \favg{} score. Instead, it provides an axis for asking a different question: how do LLM errors change as stance inference becomes more complex?

Our main finding is that LLM errors do not degrade smoothly with \sici{}. Instead, they exhibit a phase-transition-like shift between error regimes. At low complexity, models tend to over-attribute stance, with a pronounced bias toward \textsc{Against}. At intermediate complexity, accuracy drops sharply, forming an unstable boundary region. At high complexity, predictions rapidly concentrate on \textsc{None}, producing an abstention-dominated regime. Piecewise regression significantly outperforms a linear model, indicating that this pattern is not well explained as a simple monotonic decline in difficulty.

This regime structure is qualitatively consistent across models and datasets. Stronger models improve aggregate performance and move transition boundaries, but they do not remove the shift itself: GPT-4o reduces low-complexity \textsc{Against} over-prediction while becoming more prone to \textsc{None} predictions in high-complexity regions. Cross-dataset analysis further shows that \textsc{None} is not semantically uniform. In SemEval, high-\sici{} examples often contain implicit stance, so \textsc{None} predictions frequently reflect false abstention. In VAST, many high-\sici{} examples are genuinely underspecified and labeled \textsc{None}, making abstention more often appropriate.

Finally, we return to the prompting bottleneck through a systematic intervention study. We evaluate a broad set of prompting-based interventions, including chain-of-thought, multi-step reasoning, \sici{}-aware prompting, retrieval augmentation, multi-agent debate, and voting. These interventions do not reliably break the high-complexity bottleneck. Instead, they often move the model along the attribution--abstention axis: prompts that encourage indirect inference can reduce some \textsc{None} predictions but increase false attribution, whereas prompts that emphasize caution or scope clarification can reduce over-attribution while pushing the model toward excessive \textsc{None}.

Our contributions are:

(1) We identify a prompting bottleneck in LLM stance detection: hard examples are not only under-served by stronger elicitation, but organized by semantic-pragmatic inference complexity.

(2) We introduce \sici{}, a seven-dimensional diagnostic measure of stance inference complexity, and show that it explains model behavior beyond surface features such as length, lexical diversity, and negation density.

(3) We show that LLM stance errors exhibit a phase-transition-like regime shift: from low-complexity over-attribution, through an unstable boundary region, to high-complexity \textsc{None} abstention.

(4) We demonstrate that this structure is qualitatively robust across models and datasets, and that prompting, retrieval, debate, and voting interventions often trade false attribution against false abstention rather than resolving both.

\section{Related Work}

\paragraph{Stance detection.}
Stance detection has been formalized in shared tasks and benchmark datasets such as SemEval-2016 Task 6 \citep{mohammad2016semeval}, VAST \citep{AllawayM20}, P-Stance \citep{li2021p}, and multi-target or conversation-based variants \citep{wei2018multi,li2022improved,10415673}. Earlier neural approaches use target-specific attention, memory networks, graph models, and transfer learning to model the relation between a text and a target \citep{ijcai2017p557,wei2019modeling,TPDG,zhang2020enhancing}. Pretrained language models and tweet-specific encoders such as BERT and BERTweet further improve in-domain and cross-target performance \citep{devlin-etal-2019-bert,DBLP:conf/emnlp/NguyenVN20}. Recent work also incorporates background knowledge, commonsense, or prompt-based reasoning \citep{liu2021enhancing,li-etal-2023-stance,math12040568}. Our work is complementary: rather than proposing another stance classifier, we ask which instance properties systematically determine when LLM classifiers fail.

\paragraph{Zero-shot and LLM-based stance inference.}
Zero-shot stance detection is challenging because test targets may not appear during training \citep{AllawayM20,liang2022zero,allaway2021adversarial}. LLM prompting provides a natural zero-shot interface, and chain-of-thought or explanation-based variants have been explored for implicit stance and social media reasoning \citep{wei2022chain,gatto-etal-2023-chain,math12071119}. Recent LLM-era stance work has explored reasoning over ideological or tree-structured perspectives, retrieval-augmented knowledge, and stance-specific prompting for zero-shot targets \citep{zhang2024knowledge, taranukhin-etal-2024-stance,zhang-etal-2024-llm-driven}. Other recent studies emphasize multi-path reasoning for interpretability \citep{zhang-etal-2025-mprf}, LLM-driven implicit target augmentation for target-sparse examples \citep{ji-etal-2025-llm-driven}, and stereotype-sensitive evaluation of zero-shot LLM stance detection \citep{dubreuil-etal-2025-stereotypes, zhang2025logic}. We differ from these lines by treating difficulty itself as the object of measurement. Instead of adding reasoning paths, targets, or fairness probes, \sici{} asks which semantic-pragmatic properties predict when such systems fail.

\paragraph{Difficulty, calibration, and regime shifts.}
Instance difficulty is often estimated from model uncertainty, agreement, or surface features. Such signals are useful but can become circular when the same model both defines and evaluates difficulty. We instead define \sici{} using semantic-pragmatic attributes of the target--text pair, then validate its agreement across scorers and its relationship to independent prediction behavior. Our notion of a regime shift is inspired by work on emergent abilities and phase-transition-like behavior in machine learning systems \citep{wei2022emergent}. Unlike scaling-law studies, which analyze model capability as a function of model or compute scale, we study behavioral transitions as a function of \emph{instance complexity}.

\section{The SICI Framework}
\label{sec:method}

\paragraph{Seven Dimensions.}

\sici{} assigns each target--text pair seven 0--4 scores covering target visibility, scope alignment, pragmatic implicitness, knowledge need, context dependence, label ambiguity, and polarity--stance mismatch (Table~\ref{tab:sici-dims}). Details of the corresponding inference stages are given in Appendix~\ref{app:details}.

\begin{table*}[t]
\centering
\small
\begin{tabularx}{\textwidth}{p{0.15\textwidth}p{0.28\textwidth}p{0.17\textwidth}X}
\toprule
Dimension & Question measured & Low score & High-score example \\
\midrule
$V$: Target visibility & Is the target explicitly mentioned? & target named directly & stance toward a politician inferred from a policy tweet \\
$S$: Scope alignment & Is the text mainly about the target? & text is on target & text discusses a related but shifted topic \\
$P$: Pragmatic implicitness & Is stance expressed indirectly? & direct support/opposition & sarcasm, metaphor, rhetorical implication \\
$K$: Knowledge requirement & Is background knowledge needed? & self-contained text & requires knowing a law, event, or group relation \\
$C$: Context dependency & Is external conversational context needed? & standalone text & reply or quote lacking prior context \\
$A$: Label ambiguity & Is the gold label semantically contestable? & clear label boundary & mixed or underspecified stance \\
$G$: Polarity--stance gap & Does sentiment polarity align with stance? & sentiment matches stance & positive affect used to oppose via irony \\
\bottomrule
\end{tabularx}
\caption{The seven semantic-pragmatic dimensions of \sici{}. Each dimension is scored from 0 (low inference burden) to 4 (high inference burden). Together they track a multi-stage stance inference chain from target identification to polarity--stance bridging.}
\label{tab:sici-dims}
\end{table*}

\paragraph{Index Definition.}

Given seven scores $d_1,\ldots,d_7 \in \{0,1,2,3,4\}$, we define:
\begin{align}
\sici{}(x,t) =&\ 0.65 \cdot \frac{\mathrm{mean}(d_1,\ldots,d_7)}{4} \notag\\
&+0.35 \cdot \frac{\max(d_1,\ldots,d_7)}{4}.
\end{align}
The index ranges from 0 to 1. The mean term captures cumulative load, while the max term captures bottlenecks where one severe ambiguity can dominate the instance. We use equal dimension weights because a pilot comparison against a manually weighted variant produced almost identical scores (Pearson $r=0.9996$).

\paragraph{Scoring Protocol and Reliability.}

The main \sici{} scores are produced by GPT-4o-mini using dimension-specific rubrics and independent 0--4 judgments. To test whether \sici{} is a model idiosyncrasy, we additionally score a stratified sample of 200 instances with Claude Haiku 4.5 and DeepSeek-V3. Pairwise Spearman correlations are high: 0.829 (GPT vs. Claude), 0.853 (GPT vs. DeepSeek), and 0.884 (Claude vs. DeepSeek). The three-way ordinal Krippendorff's $\alpha$ is 0.771, above the conventional threshold for substantial agreement. This does not remove the need for human validation, but it supports the claim that \sici{} captures stable properties of the target--text pair rather than a single model's prediction preference.

The scoring prompt is deliberately separated from stance prediction. The scorer is asked to judge the seven attributes of the instance, not to predict \textsc{Favor}, \textsc{Against}, or \textsc{None}. This separation matters because a difficulty measure based on the same prediction confidence that later enters the evaluation would risk circularity. In our analysis, \sici{} is computed before any intervention comparison and is held fixed across all downstream models and prompts. The same \sici{} value is therefore used to analyze GPT-3.5, GPT-4o-mini, DeepSeek-V3, GPT-4o, and all intervention variants.

\paragraph{From Dimensions to Regimes.}

\sici{} supports two complementary uses. First, as a continuous score, it orders instances by expected inference complexity. Second, with empirically fitted boundaries, it partitions the data into regimes. We use two transition points, $b_1=0.45$ and $b_2=0.70$, obtained from segmented-regression analysis on SemEval. Low-complexity instances fall below $b_1$; intermediate instances occupy the boundary region where over-attribution and abstention compete; high-complexity instances exceed $b_2$ and are dominated by target invisibility, scope mismatch, and \textsc{None}-oriented abstention behavior. The thresholds are not intended as universal constants. Their role is diagnostic: they expose where a model changes its decision strategy as the target--text relation becomes less direct.

\section{Experimental Setup}
\label{sec:setup}

\paragraph{Datasets.}

We evaluate on two core benchmarks. SemEval-2016 Task 6 contains English tweets labeled \textsc{Favor}, \textsc{Against}, or \textsc{None} toward social and political targets \citep{mohammad2016semeval}. Our main SemEval analysis uses 1,249 test instances across five targets. VAST is a zero-shot stance dataset designed for unseen-topic generalization \citep{AllawayM20}; we use the unseen-topic test split with 1,460 instances. We also report supporting cross-dataset model scores on MTSD and P-Stance where available, but the central cross-dataset regime analysis uses SemEval and VAST.

\begin{table}[t]
\centering
\small
\begin{tabular}{lrrcc}
\toprule
Dataset & $N$ & Targets & Labels & Mean \sici{} \\
\midrule
SemEval & 1,249 & 5 & F/A/N & $\sim$0.35 \\
VAST unseen & 1,460 & many & F/A/N & $\sim$0.38 \\
MTSD & 500 & 2 & F/A/N & $\sim$0.33 \\
P-Stance & 777 & 1 & F/A & 0.267 \\
\bottomrule
\end{tabular}
\caption{Datasets used in the analysis. F/A/N denotes \textsc{Favor}, \textsc{Against}, and \textsc{None}.}
\label{tab:datasets}
\end{table}

\paragraph{Models and Metrics.}

We evaluate GPT-3.5-turbo, GPT-4o-mini, DeepSeek-V3, and GPT-4o under zero-shot prompting. SemEval and P-Stance use \favg{} $=(F1_{\textsc{Favor}}+F1_{\textsc{Against}})/2$, the official SemEval-style metric that excludes \textsc{None} from the average. MTSD and VAST use macro-F1 or accuracy depending on the analysis; for cross-dataset phase analysis we focus on accuracy within \sici{} regions and separately inspect \textsc{None} label composition to avoid inflated high-\sici{} scores.

We report both instance-level and binned statistics. Instance-level correlations test whether higher \sici{} predicts correctness at the sample level. Binned correlations test whether the aggregate trend is monotonic after grouping instances into SICI intervals. This distinction is important for stance detection because individual labels are noisy, while regime-level behavior can still be stable. For SemEval, we also report \favg{} because it is the official metric and prevents a high \textsc{None} prior from artificially improving the score.

We also compute two diagnostic quantities. The first is the systematic-bias rate: the proportion of examples where the model predicts \textsc{Against} when the gold label is not \textsc{Against}, plus the proportion where it predicts \textsc{None} when the gold label is not \textsc{None}. The second is BII, a bias--SICI interaction index that compares confidence and correctness within \sici{} bins. BII is intended to capture cases where the model is not merely wrong, but wrong with high confidence.

\paragraph{Surface Baselines.}

To test whether \sici{} adds information beyond simple proxies, we compare against text length, target visibility as lexical target coverage, type-token ratio, and negation density. These baselines represent common surface-level difficulty hypotheses: longer texts may contain more evidence, low target mention rate may make inference harder, lexical diversity may increase complexity, and negation may confuse polarity-based decisions.

\paragraph{Interventions.}

For the high-complexity SemEval subset ($\sici{} \ge 0.70$, $N=187$), we evaluate 15 inference-time interventions: self-consistency, few-shot prompting, self-reflection, counterfactual reasoning, target decomposition, generated knowledge, multi-agent debate, evidence chaining, SICI-neighbor few-shot retrieval, dimension-targeted routing, Wikipedia RAG, RAG plus scope routing, cultural-camp debate, and cultural-camp debate with RAG. This set spans prompting, retrieval, routing, and debate-style methods.

The intervention suite is designed to distinguish three possible explanations for high-\sici{} failure. If the bottleneck is unstable decoding, self-consistency should help. If it is missing reasoning, chain-of-thought, reflection, counterfactual analysis, or target decomposition should help. If it is missing information, few-shot retrieval or Wikipedia RAG should help. Failure across all three families would instead support the stronger interpretation that many Phase-3 examples are underdetermined by the available text-target pair.

\section{Results}
\label{sec:results}

\paragraph{\sici{} Predicts Accuracy Better Than Surface Proxies.}

On the merged non-\textsc{None} SemEval+VAST set ($N=1{,}960$), \sici{} has the strongest and directionally correct relationship with accuracy. At the sample level, the point-biserial correlation between \sici{} and correctness is $r=-0.2405$ ($p=3.4\times10^{-27}$). At the binned level, the Spearman correlation is $r=-0.9515$ ($p=2.3\times10^{-5}$). Table~\ref{tab:baseline} shows that surface baselines are either weaker or directionally misleading. Target visibility has a high positive binned correlation because explicit targets are easier, but it captures only one dimension and cannot explain high-\sici{} collapse caused by scope and pragmatic ambiguity.

\begin{table}[t]
\centering
\small
\begin{tabular}{lrr}
\toprule
Metric & Sample $r$ & Binned $\rho$ \\
\midrule
\sici{} & \textbf{-0.2405} & \textbf{-0.9515} \\
Target visibility & +0.1328 & +0.9245 \\
Text length & +0.0648 & +0.6991 \\
Type-token ratio & -0.0508 & -0.4877 \\
Negation density & -0.0174 & +0.0813 \\
\bottomrule
\end{tabular}
\caption{Correlation with correctness on merged non-\textsc{None} SemEval+VAST instances. Negative values mean that higher difficulty predicts lower accuracy.}
\label{tab:baseline}
\end{table}

\paragraph{A Dual-Fixation Regime Shift.}

\begin{figure*}[t]
\centering
\begin{subfigure}{0.49\textwidth}
  \centering
  \includegraphics[width=\linewidth]{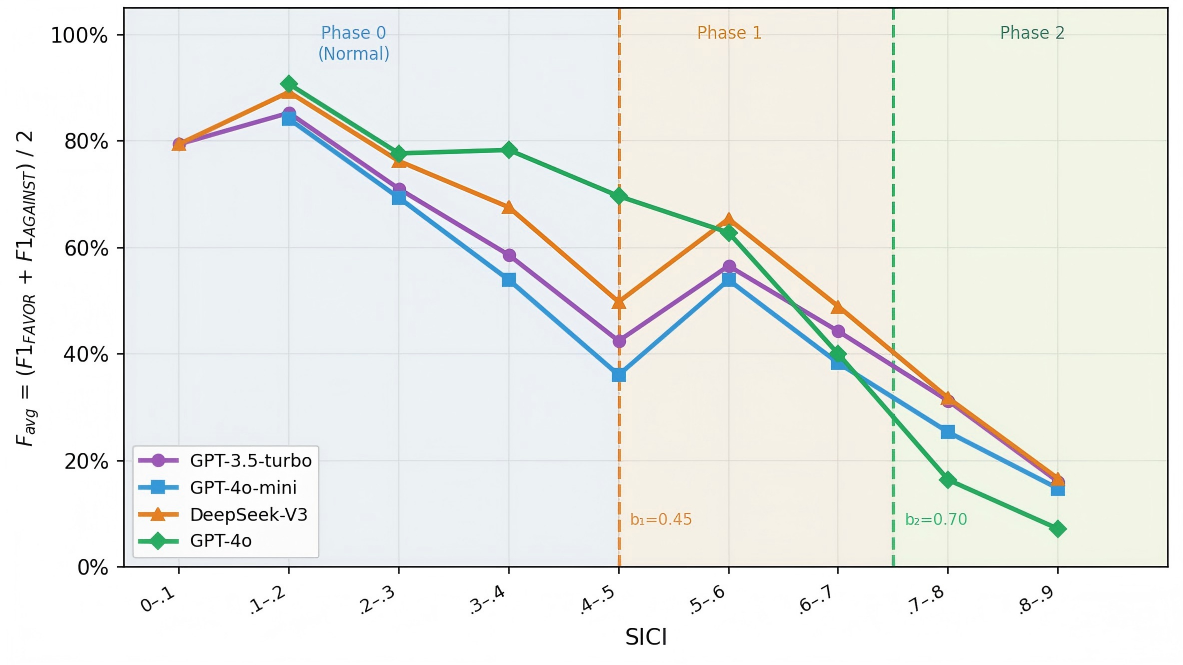}
  \caption{Official SemEval \favg{} across four models.}
\end{subfigure}
\hfill
\begin{subfigure}{0.49\textwidth}
  \centering
  \includegraphics[width=\linewidth]{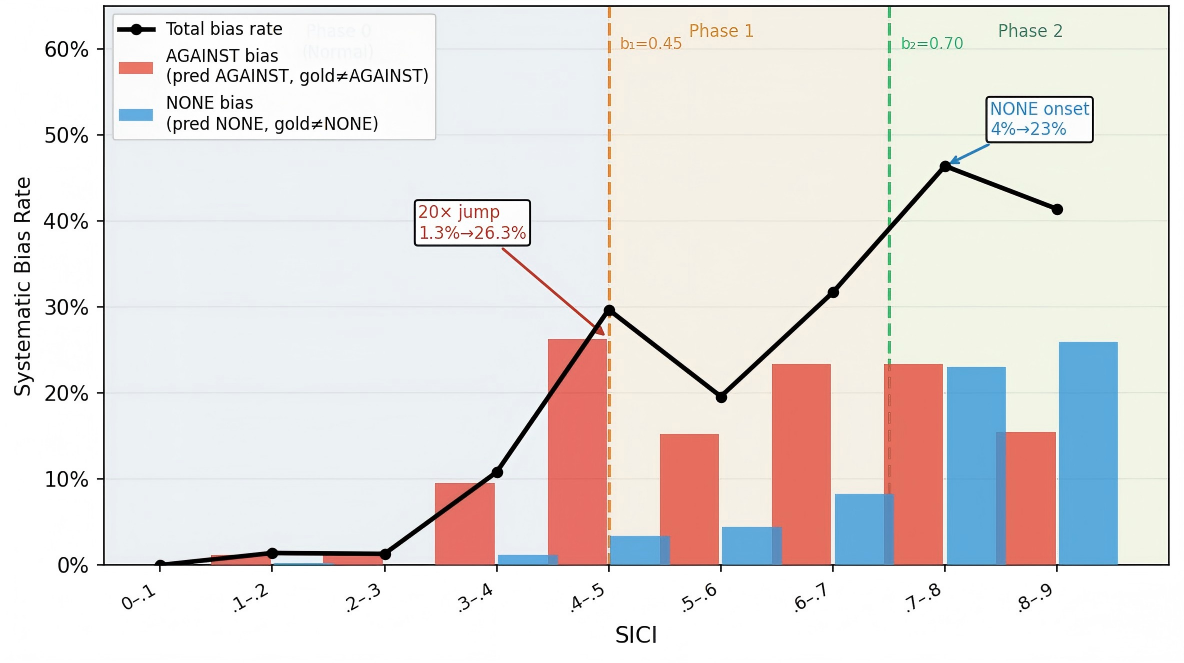}
  \caption{Systematic prediction-bias rate for GPT-4o-mini.}
\end{subfigure}
\caption{User-provided main regime-shift figures. As \sici{} increases, stance prediction quality declines across models, while systematic bias reveals two critical jumps: \textsc{Against} fixation near $b_1=0.45$ and \textsc{None} escape near $b_2=0.70$.}
\label{fig:phase}
\end{figure*}

Figure~\ref{fig:phase} summarizes the core phenomenon. As \sici{} increases, model performance decreases, but the more diagnostic signal is the systematic prediction-bias rate. Lower-complexity errors are dominated by \textsc{Against} over-prediction. Around \sici{} $\approx 0.45$, accuracy reaches a trough: \textsc{Against} fixation begins to fail, while \textsc{None} escape has not yet become dominant. Above roughly 0.70, \textsc{None} predictions rise sharply. A segmented-regression comparison confirms that the three-regime structure fits better than a linear trend ($F(2,14)=16.82$, $p=1.89\times10^{-4}$).

The fit improvement is large in absolute terms. On 18 SICI bins, the residual sum of squares drops from 0.2141 for a single linear model to 0.0629 for the two-breakpoint segmented model, a 70.6\% reduction. The first breakpoint is stable around 0.45 across GPT-3.5-turbo, GPT-4o-mini, and DeepSeek-V3 (0.425--0.450). The second breakpoint varies more (0.700--0.800), suggesting that models differ mainly in when they begin to rely on \textsc{None} escape.

This pattern motivates an attribution--abstention account of the failure. The model is not simply uncertain. It falls back on different heuristics in different regions: first a tendency to treat stance-bearing social media text as opposition, then a tendency to abstain with \textsc{None} when the target--text relation becomes indirect.

Table~\ref{tab:bintrajectory} makes the shift concrete. The low-\sici{} bins contain many \textsc{Against} predictions and high accuracy. The boundary bin at 0.4--0.5 is the weakest point: accuracy drops to 42.3\%, while \textsc{Against} remains common and \textsc{None} has not yet become the dominant output. After 0.7, the model's output distribution changes qualitatively. \textsc{None} predictions rise from 54.5\% in the 0.7--0.8 bin to 75.0\% above 0.8, while \textsc{Favor} nearly disappears. This is why a single monotone ``harder means lower accuracy'' story is incomplete: the same increasing complexity first produces over-commitment to opposition and then over-abstention.

\begin{table}[t]
\centering
\small
\begin{tabular}{lrrrr}
\toprule
\sici{} bin & $N$ & \%Against & \%None & Acc. \\
\midrule
0.0--0.1 & 24 & 4.2 & 0.0 & 87.5 \\
0.1--0.2 & 599 & 61.1 & 1.2 & 85.8 \\
0.2--0.3 & 158 & 61.4 & 1.9 & 71.5 \\
0.5--0.6 & 250 & 70.0 & 16.8 & 57.2 \\
0.6--0.7 & 337 & 66.8 & 29.4 & 52.2 \\
0.7--0.8 & 235 & 43.4 & 54.5 & 38.3 \\
0.8--1.0 & 64 & 23.4 & 75.0 & 37.5 \\
\bottomrule
\end{tabular}
\caption{SemEval GPT-4o-mini prediction trajectory by \sici{} bin. Percent columns are model prediction rates, not gold label rates.}
\label{tab:bintrajectory}
\end{table}

\begin{figure*}[t]
\centering
\begin{subfigure}{0.49\textwidth}
  \centering
  \includegraphics[width=\linewidth]{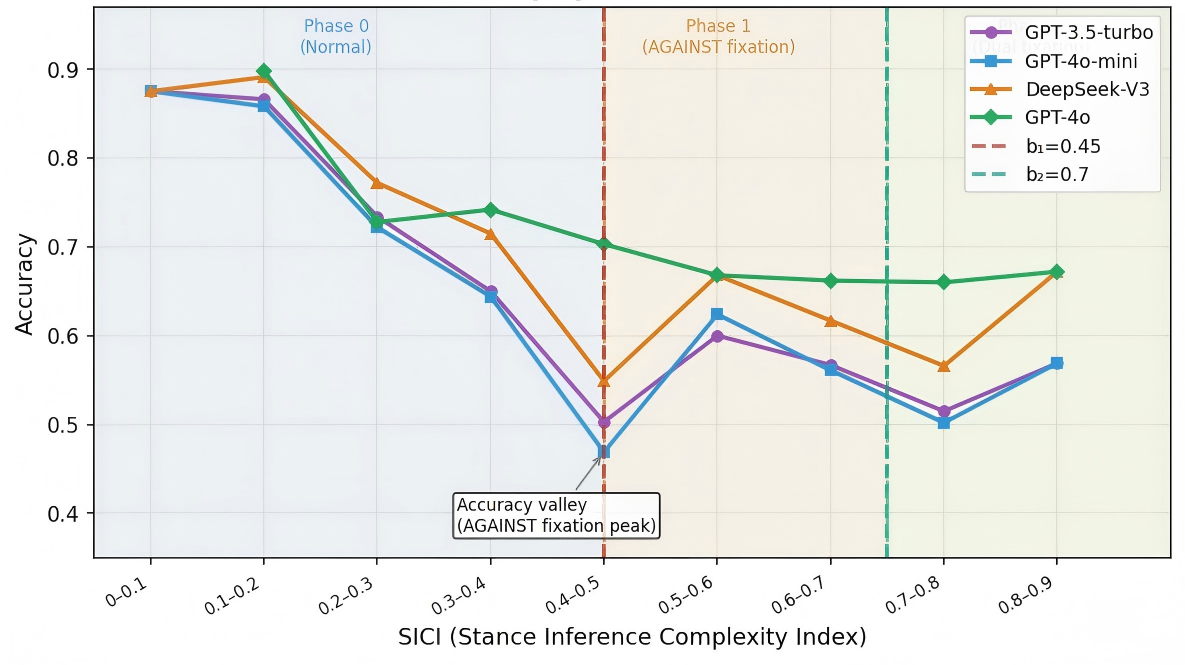}
  \caption{Accuracy by \sici{} bin across four models.}
\end{subfigure}
\hfill
\begin{subfigure}{0.49\textwidth}
  \centering
  \includegraphics[width=\linewidth]{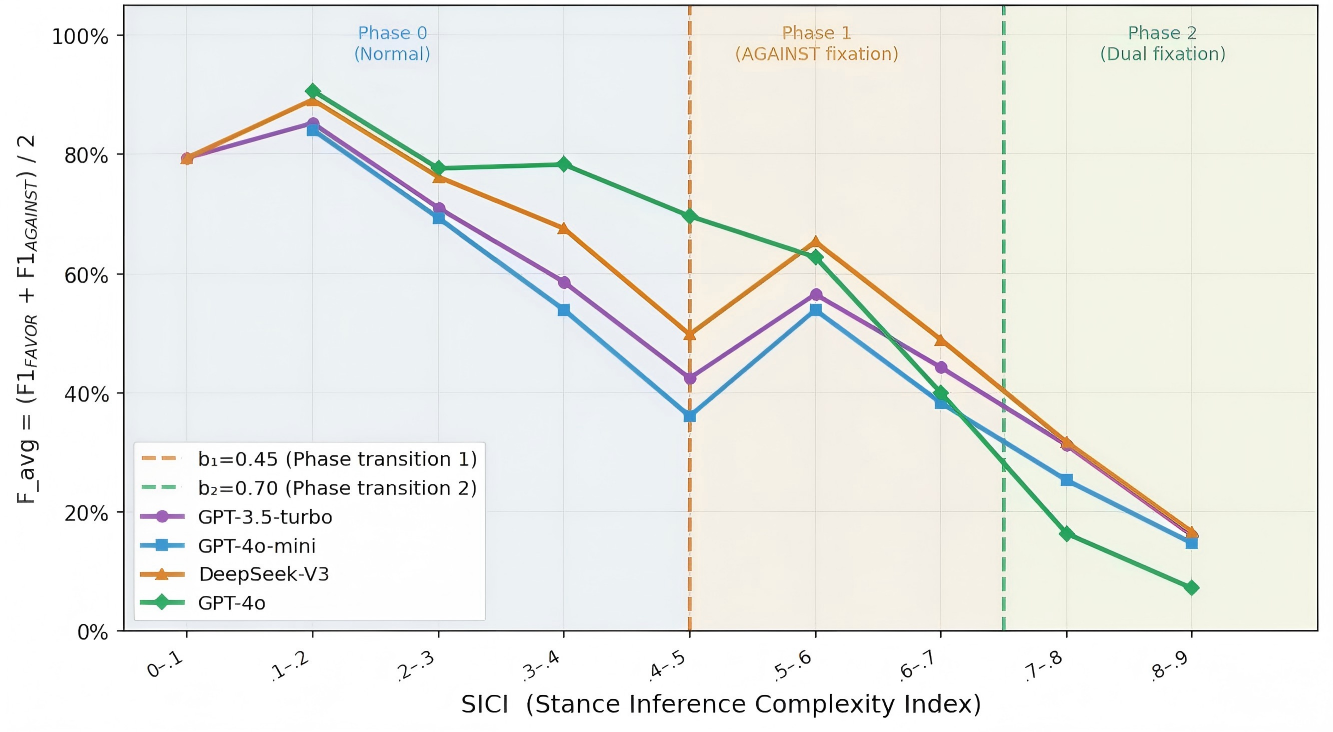}
  \caption{Alternative \favg{} view excluding \textsc{None}.}
\end{subfigure}
\caption{Additional user-provided visualizations of the same regime structure. Accuracy alone can obscure the transition because high-\sici{} bins often contain more \textsc{None} labels; \favg{} and systematic-bias rate isolate the stance-inference failure more directly.}
\label{fig:phase-extra}
\end{figure*}

Figure~\ref{fig:phase-extra} shows two complementary views. Raw accuracy exhibits a partial rebound after the first boundary because \textsc{None} becomes more frequent in some high-\sici{} bins. This rebound is misleading if interpreted as easier inference: a model can appear more accurate simply because the label prior has shifted toward \textsc{None}. The official \favg{} view and the systematic-bias view avoid this artifact. The former ignores \textsc{None} when computing SemEval quality; the latter directly measures when the model predicts \textsc{Against} or \textsc{None} against the gold label. Together, the four views show why the regime-shift claim is not a visual artifact of one metric.

\paragraph{Stronger Models Move Boundaries but Do Not Remove the Regime.}

GPT-4o substantially improves overall performance compared with GPT-4o-mini, but the \sici{} relationship remains significant. Table~\ref{tab:gpt4o} shows that GPT-4o reduces the lower-complexity \textsc{Against} peak and improves the 0.4--0.5 boundary region, but its high-\sici{} \textsc{None} rate is even higher. Thus scale or capability changes the failure profile; it does not erase complexity-conditioned behavior.

The cross-model pattern also clarifies what ``stronger'' means in this setting. GPT-4o is better at recovering explicit and moderately implicit stances, but its improvement is not equivalent to solving the inference chain. The first breakpoint moves left, which means the model exits the \textsc{Against}-fixation regime earlier. The second breakpoint moves right, which means the model postpones its strongest \textsc{None} escape. Yet the correlation remains negative, and the high-\sici{} region is still governed by abstention behavior. Capability therefore stretches the phase boundaries rather than flattening the phase diagram.

\begin{table}[t]
\centering
\small
\begin{tabular}{lrr}
\toprule
Measure & GPT-4o-mini & GPT-4o \\
\midrule
Overall macro-F1 & $\sim$0.61 & 0.737 \\
\textsc{Against} peak (0.3--0.4) & 81.4 & 51.4 \\
Accuracy at 0.4--0.5 & 42.3 & 70.3 \\
\textsc{None} rate at 0.7--0.8 & 54.5 & 84.3 \\
First boundary $b_1$ & 0.450 & 0.250 \\
Second boundary $b_2$ & 0.700 & 0.825 \\
\sici{}--accuracy $\rho$ & -0.243 & -0.191 \\
\bottomrule
\end{tabular}
\caption{Stronger models shift the regime boundaries but preserve a significant relationship between \sici{} and accuracy. Percent-valued rows omit percent signs for compactness.}
\label{tab:gpt4o}
\end{table}

\paragraph{Cross-Dataset Validation Separates Two Kinds of \textsc{None}.}

VAST provides an important contrast. Its high-\sici{} region has high overall accuracy because the gold labels are mostly \textsc{None}: 250 of 288 Phase-3 examples (86.8\%). This could appear to contradict the SemEval high-\sici{} failure pattern. However, after separating non-\textsc{None} cases, the difficulty remains: VAST Phase-3 non-\textsc{None} accuracy is 0.579, and merged SemEval+VAST non-\textsc{None} accuracy drops monotonically from 0.755 to 0.605 to 0.323 across the three \sici{} phases.

\begin{table}[t]
\centering
\small
\begin{tabular}{lrr}
\toprule
Phase & $N$ & Accuracy \\
\midrule
\sici{} $<0.45$ & 1,069 & 0.755 \\
0.45 $\le$ \sici{} $<0.70$ & 760 & 0.605 \\
\sici{} $\ge 0.70$ & 131 & 0.323 \\
\bottomrule
\end{tabular}
\caption{Merged SemEval+VAST non-\textsc{None} accuracy by \sici{} phase.}
\label{tab:mergedphase}
\end{table}

This comparison reveals a theoretical distinction. In SemEval Phase 3, many \textsc{Against} instances are incorrectly mapped to \textsc{None}: 68 of 87 \textsc{Against} cases become \textsc{None}. In VAST Phase 3, \textsc{None} is often a reasonable abstention because the target--text relation is genuinely underspecified. \sici{} therefore separates two superficially similar outputs: \emph{failed inference fixation} and \emph{information-insufficient abstention}.

Additional dimension-level analysis, reported in Appendix~\ref{app:dimension-routing}, shows that target visibility and scope alignment are the strongest individual drivers, while the remaining dimensions help distinguish target sparsity from genuinely complex stance inference.

\paragraph{Inference-Time Interventions Hit a Ceiling.}

Table~\ref{tab:interventions} shows the high-\sici{} intervention results. The best methods are still zero-shot and self-consistency at 58.3\%. Wikipedia RAG comes closest at 57.8\%, showing that real external knowledge is somewhat better than generated knowledge but still does not exceed the baseline. SICI-guided retrieval and dimension-targeted routing underperform zero-shot. Multi-agent debate is harmful, and cultural-camp debate collapses by over-predicting \textsc{Favor}.

\begin{table}[t]
\centering
\small
\begin{tabular}{lrr}
\toprule
Method & Acc. & $\Delta$ vs. ZS \\
\midrule
Zero-shot & 58.3 & -- \\
Self-consistency & 58.3 & 0.0 \\
Wikipedia RAG & 57.8 & -0.5 \\
Target decomposition & 55.1 & -3.2 \\
Dimension routing & 52.9 & -5.4 \\
Self-reflection & 50.8 & -7.5 \\
SICI retrieval & 49.2 & -9.1 \\
Few-shot & 47.1 & -11.2 \\
Multi-agent debate & 47.1 & -11.2 \\
Cultural-camp debate & 31.0 & -27.3 \\
\bottomrule
\end{tabular}
\caption{Representative intervention results on $\sici{}\ge0.70$ SemEval examples ($N=187$).}
\label{tab:interventions}
\end{table}

These failures clarify the nature of the ceiling. Additional reasoning does not help when the missing ingredient is not a reasoning step but an underdetermined pragmatic link between the text and target. Retrieval helps only marginally because the most difficult cases often require conversational, author-specific, or discourse-level context that generic Wikipedia knowledge cannot supply.

The failures also reveal a label-prior pendulum. Some interventions, such as indirect inference and scope clarification, reduce \textsc{Against} errors but over-correct toward \textsc{None}. In the original local analysis, indirect stance inference predicts \textsc{None} for 232 of 299 triggered cases (77.6\%), although the gold \textsc{None} rate is only 56\%. Scope clarification is more extreme: 142 of 154 triggered cases are predicted as \textsc{None} (92.2\%), while the gold \textsc{None} rate is 45.5\%. Debate-style prompting moves in the opposite direction in some variants. Cultural-camp debate predicts \textsc{Favor} 105 times even though only five gold examples are \textsc{Favor}, and its RAG-augmented variant preserves almost the same skew. These shifts can improve global macro-F1 slightly when they counteract a dataset-level bias, but they do not solve the local high-\sici{} decision. A useful intervention for stance detection must therefore do more than change a model's prior over labels; it must calibrate the strength of the target--text link.

Appendix~\ref{app:dimension-routing} further reports routing and confidence analyses: \sici{}-guided routes can improve global macro-F1 by shifting label priors, but they do not solve the local high-\sici{} subset, and high complexity is not reducible to low confidence.

\section{Discussion}
\label{sec:discussion}

\paragraph{\sici{} as a diagnostic rather than a leaderboard metric.}
\sici{} should not replace task metrics. Instead, it complements them by identifying where a model's aggregate score comes from. Two models with similar macro-F1 may differ in whether they fail through \textsc{Against} fixation, \textsc{None} escape, or boundary-region confusion. Reporting performance stratified by \sici{} would make stance evaluation more informative.

This diagnostic role is especially important for prompt-based evaluation. A low score can arise because the model lacks task competence, because the prompt induces the wrong label prior, or because the instance itself provides too little target-conditioned evidence. \sici{} helps separate these cases by making the source of difficulty inspectable: the target may be invisible, the text may be off-scope, the stance may be pragmatic, external knowledge may be required, or the gold label may be intrinsically ambiguous. This turns error analysis from a post-hoc list of mistakes into a structured account of where the target--text inference chain breaks.

\paragraph{Why high-complexity prompting fails.}
The intervention results suggest a practical warning for LLM-based stance systems. Prompting strategies often shift label priors rather than improve fine-grained calibration. Indirect-inference and scope-clarification prompts reduce one kind of error but overshoot toward \textsc{None}; debate prompts can introduce new hallucinated stances. This explains the ``pendulum'' effect observed in our experiments: interventions swing the model from one fixation mode to another.

The failure is therefore not simply that the tested prompts are weak. Many interventions add exactly the resources that prompt engineering usually assumes to be helpful: explicit reasoning, multiple samples, retrieved knowledge, or disagreement among agents. Their limited effect suggests that the hard cases often require information that is not recoverable from the visible text-target pair, or require pragmatic commitments that the model cannot license without over-interpreting the author. In such cases, a better prompt may only choose a different trade-off between false attribution and false abstention.

\paragraph{Toward model-level and context-level solutions.}
If high-\sici{} failures are caused by missing pragmatic context or ambiguous target--text relations, inference-time prompting may be insufficient. More promising directions include SICI-stratified training, calibrated abstention policies, retrieval of conversational context rather than encyclopedic background, and human-in-the-loop treatment of intrinsically ambiguous cases.

\paragraph{Implications for benchmark construction.}
The analysis also suggests that stance benchmarks should report their complexity distribution. A dataset dominated by low-\sici{} examples primarily tests direct target matching and sentiment-to-stance mapping. A dataset with many high-\sici{} examples tests pragmatic inference, context recovery, and abstention calibration. Without this distribution, two datasets with the same label set can evaluate very different skills. Reporting \sici{} histograms and phase-stratified scores would make cross-dataset comparisons less dependent on hidden differences in target visibility and topic alignment.

The per-target results from the original analysis illustrate this point. In VAST, the \emph{election} target is especially difficult: all evaluated models are near 26--34\% macro-F1, close to random-level behavior for a three-way task. By contrast, P-Stance removes the \textsc{None} label and produces much higher scores for stronger models. These differences are not just dataset names; they reflect distinct inference regimes. A hard benchmark for stance detection should therefore be constructed from high-\sici{} non-\textsc{None} examples rather than from examples that are merely label-balanced.

\paragraph{Implications for LLM evaluation.}
For LLMs, the most concerning failures are not always low-confidence errors. In our analysis, high-\sici{} examples often elicit confident but wrong predictions, especially around the boundary where \textsc{Against} fixation and \textsc{None} escape compete. This behavior matters for downstream use: a stance system that returns a single label without exposing instance complexity may be least reliable exactly when its output looks decisive. \sici{} can therefore serve as a triage signal: low-complexity predictions may be used normally, intermediate cases may require calibration checks, and high-complexity cases should trigger abstention, context retrieval, or human review.

\section{Conclusion}

We introduced \sici{}, a semantic-pragmatic complexity index for diagnosing LLM stance detection. Across datasets and models, \sici{} predicts where LLMs fail and reveals an attribution--abstention regime shift from low-complexity over-prediction to high-complexity \textsc{None} escape. Stronger models move the transition boundaries but do not eliminate the phenomenon, and prompting, retrieval, debate, and voting interventions often shift the model's operating point rather than remove the high-complexity bottleneck. The main implication is that stance detection errors should be analyzed as structured failures of target-conditioned inference rather than as undifferentiated classification mistakes. Future stance systems should therefore combine model improvements with complexity-aware evaluation, calibrated abstention, and richer context acquisition.

\newpage

\section*{Limitations}

First, the main \sici{} scores are LLM-generated. Although three independent LLM scorers show substantial agreement, human annotation is needed to fully validate the scale and rule out shared model biases. Second, the experiments focus on English social-media datasets; long-form, multilingual, and multimodal stance settings may exhibit different dimension weights. Third, the high-\sici{} non-\textsc{None} subset is relatively small, especially in VAST, so future work should construct larger hard-instance benchmarks. Finally, our intervention study is inference-time only; it does not test whether supervised fine-tuning on SICI-stratified data can reduce the high-complexity ceiling.

\section*{Ethical Considerations}

Stance detection can be used for beneficial analysis of public discourse, but also for political profiling, surveillance, and manipulation. Our work is diagnostic and does not release a new classifier intended for deployment. We use publicly available benchmark datasets and API-accessible LLMs under their respective terms of use, and we do not redistribute dataset contents or model weights. Because high-\sici{} instances are often ambiguous or context-dependent, automated decisions on such cases should not be treated as ground truth. Systems using stance predictions in sensitive settings should report uncertainty, allow human review, and avoid inferring personal beliefs from sparse or indirect text.

\bibliography{main_refs}

\appendix

\section{Additional Experimental Details}
\label{app:details}

\paragraph{Inference-chain interpretation.}
The seven \sici{} dimensions can be read as a multi-stage stance inference chain. A model must identify the target ($V$), decide whether the text's scope bears on that target ($S$), decode pragmatic cues ($P$), retrieve needed knowledge ($K$), integrate missing context ($C$), resolve label ambiguity ($A$), and bridge sentiment polarity to stance ($G$). A severe failure at any stage can derail the final label.

\paragraph{Phase definitions.}
We use $b_1=0.45$ and $b_2=0.70$ as empirical transition points, yielding Phase 1 ($\sici{}<0.45$), Phase 2 ($0.45\le\sici{}<0.70$), and Phase 3 ($\sici{}\ge0.70$). These thresholds are selected from the SemEval segmented-regression analysis and then reused for VAST.

\paragraph{Full intervention set.}
The full Phase-3 intervention suite includes zero-shot, self-consistency, few-shot, self-reflection, counterfactual reasoning, target decomposition, generated knowledge, standard multi-agent debate, evidence chaining, SICI-nearest-neighbor few-shot retrieval, dimension-targeted adaptive prompting, Wikipedia RAG, RAG plus scope routing, cultural-camp debate, and cultural-camp debate plus RAG. Evidence chaining is excluded from substantive comparisons because a formatting failure caused label parsing to collapse.

\section{Additional Dimension, Routing, and Confidence Analyses}
\label{app:dimension-routing}

\paragraph{Dimension-level gradients.}
Single-dimension analysis on SemEval indicates that target visibility and scope alignment are the strongest drivers. Accuracy drops from 0.909 to 0.607 between $V=0$ and $V=4$, and from 0.886 to 0.627 between $S=0$ and $S=4$. Their interaction is especially diagnostic: in the SemEval $V=4,S=4$ cell ($N=247$), the model predicts \textsc{None} for 85.8\% of examples while the gold \textsc{None} rate is only 50.2\%. In VAST, the same $V=4,S=4$ pattern corresponds to a mostly valid \textsc{None} abstention: \textsc{None}-prediction rate 95.5\% and gold \textsc{None} rate 96.6\%. This suggests that the same inference-chain break can either expose a dataset's legitimate underspecification or reveal a model's erroneous escape behavior.

The remaining dimensions help explain why target visibility alone is insufficient. Pragmatic implicitness captures cases where the target is visible but the stance is conveyed through irony, metaphor, or indirect evaluation. Knowledge requirement captures cases where a stance depends on knowing an event, policy, or social relation. Label ambiguity captures instances where the text may support multiple defensible interpretations. Polarity--stance gap captures the familiar failure of treating sentiment as stance. These dimensions are individually weaker than $V$ and $S$ in our SemEval analysis, but they are essential for distinguishing a merely target-sparse example from a genuinely complex stance inference problem.

Ablation results add a second perspective. Removing $S$ or $P$ changes the binned correlation most strongly, indicating that scope alignment and pragmatic implicitness are central to the monotonic regime signal. Removing $G$ slightly improves the correlation in our current equal-weight setting, which suggests that polarity--stance gap is noisier than the other dimensions on these datasets. We keep $G$ because the construct is theoretically important for stance detection, but the result indicates that future versions of \sici{} should refine how affective polarity is separated from target-conditioned stance.

\begin{table}[h]
\centering
\small
\begin{tabular}{lrrr}
\toprule
Dimension & Low acc. & High acc. & Drop \\
\midrule
$V$: target visibility & 0.909 & 0.607 & -0.302 \\
$S$: scope alignment & 0.886 & 0.627 & -0.258 \\
$K$: knowledge need & 0.769 & 0.549 & -0.220 \\
$C$: context dependence & 0.746 & 0.600 & -0.146 \\
\bottomrule
\end{tabular}
\caption{Representative SemEval single-dimension gradients. ``Low'' and ``high'' refer to low- versus high-complexity levels for that dimension; exact sample sizes vary by dimension.}
\label{tab:dimension-gradient}
\end{table}

\paragraph{SICI-guided routing.}
The original experiments also tested staged routing on the full 2,899-instance pool. A simple CoT route triggered on $\sici{}>0.6$ improves global macro-F1 from 0.5916 to 0.6132 (+2.16pp). A more targeted ISI+SC route, which applies indirect stance inference to high-$V$ cases and scope clarification to high-$S$ cases, reaches 0.6209 (+2.93pp). The best full routing policy combines ISI, SC, and CoT on 790 triggered examples (27.2\% of the pool), reaching 0.6296 (+3.80pp).

This result prevents an overly pessimistic reading of the intervention study. \sici{} can guide useful global routing: when the model's dominant error is an overactive \textsc{Against} prior, shifting some examples toward more cautious reasoning can improve macro-F1. However, the same mechanism does not solve the high-\sici{} subset. Local ISI and SC scores fall below their local baselines because they overshoot toward \textsc{None}. Thus \sici{} is valuable for diagnosis and routing, but routing is not equivalent to resolving the semantic-pragmatic ceiling.

This global-local gap is useful diagnostically. A routing policy can improve the dataset-level score by correcting the most common prior error, but its gains depend on the distribution of phases in the evaluation set. In a deployment setting where high-\sici{} non-\textsc{None} cases are the main concern, the same routing rule could be harmful. We therefore treat routing as an operational use of \sici{}, not as evidence that the underlying semantic-pragmatic ambiguity has been removed.

\paragraph{Confidence failure.}
High-\sici{} examples are not simply low-confidence cases. BII is negatively correlated with \sici{} ($\rho=-0.308$), indicating that the gap between confidence and correctness worsens as complexity rises. The most problematic region is the transition band, where \textsc{Against} fixation has become unreliable but \textsc{None} escape has not yet become a calibrated abstention strategy. In this region, the model can select a wrong label with apparent confidence.

This finding changes how we interpret \textsc{None}. A well-calibrated model should reserve \textsc{None} for cases where the text does not support a stance toward the target. In the high-\sici{} SemEval subset, \textsc{None} often functions instead as an uncalibrated failure mode: the model gives up on implicit \textsc{Against} examples even when the gold label is not \textsc{None}. In VAST, by contrast, high-\sici{} \textsc{None} is often a valid response. Confidence alone cannot distinguish these two cases; the target--text complexity structure is needed.

\end{document}